%% file: main.tex
\documentclass[a4paper,12pt]{article}
\pdfoutput=1
\usepackage{amssymb}
\usepackage[utf8]{inputenc}
\usepackage[style=ieee]{biblatex}
\usepackage{float}
\usepackage{graphicx}
\usepackage[margin=1in]{geometry}
\usepackage{hyperref}
\usepackage[normalem]{ulem}
\hypersetup{
    colorlinks=true,
    linkcolor=black,
    citecolor=cyan,
    urlcolor=cyan
}
\graphicspath{ {./images/} }
\usepackage{eurosym}
\usepackage{pdfpages}
\usepackage{authblk}
\usepackage[font=small,labelfont=bf]{caption}
\usepackage{svg}
\usepackage{multirow}
\usepackage{array}
\usepackage{amsmath}
\usepackage{subcaption}

\title{GRLib: An Open-Source Hand Gesture Detection and Recognition Python Library}
\author{Jan Warchocki$^\dag$, Mikhail Vlasenko$^\dag$, Yke Bauke Eisma}
\date{October 2023}

\begin{document}

\maketitle

\def\thefootnote{$\dag$}\footnotetext{These authors contributed equally to this work}

\input{sections/abstract}

\section{Introduction}

\input{sections/introduction}

\section{Static Gesture Methodology}
\label{sec:static_methodology}
\input{sections/static_methodology}

\section{Dynamic Gesture Methodology}
\label{sec:dynamic_methodology}
\input{sections/dynamic_meth}





\input{sections/experiments}

\input{sections/conclusion_future_work}

\printbibliography

\end{document}

%% file: sections/abstract.tex
\begin{abstract}
    Hand gesture recognition systems provide a natural way for humans to interact with computer systems. Although various algorithms have been designed for this task, a host of external conditions, such as poor lighting or distance from the camera, make it difficult to create an algorithm that performs well across a range of environments. In this work, we present GRLib: an open-source Python library able to detect and classify static and dynamic hand gestures. Moreover, the library can be trained on existing data for improved classification robustness. The proposed solution utilizes a feed from an RGB camera. The retrieved frames are then subjected to data augmentation and passed on to \textit{MediaPipe Hands} to perform hand landmark detection. The landmarks are then classified into their respective gesture class. The library supports dynamic hand gestures through trajectories and keyframe extraction. It was found that the library outperforms another publicly available HGR system - MediaPipe Solutions, on three diverse, real-world datasets. The library is available at \url{https://github.com/mikhail-vlasenko/grlib} and can be installed with \texttt{pip}.
\end{abstract}

%% file: sections/introduction.tex
In recent times, gesture recognition algorithms have provided novel ways for interaction with computer systems. Hand gesture recognition (HGR) systems are of special interest, as they allow for human-computer interactions \cite{cv_hgr} based on naturalistic gesture inputs. These systems focus exclusively on the shape and movement of the hands. Hand gesture recognition algorithms have already been used in different tasks, including robot control and entertainment systems \cite{vision_based_review}. 

Different techniques have been employed to develop HGR systems. For example, the first HGR algorithms, such as the MIT-LED glove \cite{mit_led_mention}, were glove-based systems and required the user to wear a special glove with sensors attached to it \cite{history_hgr, glove_based_systems}. However, due to user comfort and ease of use, vision-based HGR systems are currently becoming increasingly popular \cite{vision_based_review}. Approaches towards developing and improving vision-based systems range from skin colour detection \cite{colour_based_comparison, colour_based_survey} to deep learning \cite{cnn_hgr, rcnn_hgr, vision_based_review}. While vision-based systems are already frequently employed in HGR systems, they present challenges, especially when utilizing cameras that produce lower-quality images like webcams.  In \cite{difficulties_vision_based_systems} authors note that vision-based systems are often constrained by aspects such as lighting sensitivity and distance to the camera. Authors of \cite{robust_real_time} and \cite{robust_kinect} also mention complex background as a potential hindrance to accurate and robust gesture detection. 

These obstacles make it difficult to create a single algorithm that has robust gesture detection capabilities in a wide range of use cases. As a result, there is a lack of open-source, public libraries that allow their users to define their own vision-based hand gestures, and then subsequently detect and recognize them \cite{gesture_control_computer_systems}. Such a library would be of interest as it streamlines the design of systems that are controlled through hand gestures. Although constructing a robust HGR system from scratch is difficult, in recent years, there have been many advancements in this area that make designing such a system more feasible. One such advancement, \textit{MediaPipe Hands} \cite{mediapipe}, can be used to construct a real-time and accurate palm detection model, which makes an open-source hand gesture detection and recognition library possible. For instance, in \cite{paper_realtime_using_mp} authors successfully use \textit{MediaPipe Hands} to build a real-time HGR system, able to detect a set of predefined static gestures from an RGB camera \cite{paper_realtime_using_mp}.

Krishna and Sinha \cite{gesture_control_computer_systems} proposed \textit{Gestop} - a library for gesture control of computer systems. The designed algorithm utilizes \textit{MediaPipe Hands} to detect hand landmarks which are then passed to a pre-trained neural network. The library has been shown to work well on the designed task, obtaining an accuracy of 99.12\% for static, and 85\% for dynamic gestures on their validation dataset. Additionally, the library allows its users to define their own gestures through a camera live feed. The system, however, does not support existing datasets as a source for the training data, which limits the flexibility of the library. A similar approach to the task of building an HGR system has been recently used in MediaPipe Solutions \cite{mediapipe_solutions}. Similarly to \textit{Gestop}, this solution also uses \textit{MediaPipe Hands} to build the landmarks that are then passed to a neural network to perform gesture classification. This system also allows the user to define their own hand gestures. This library however, does not support dynamic gestures.

In this paper, we present a hand gesture recognition library, that allows its users to define their own gestures. More important, we aim for the library to be effective with few training samples and to be usable with lower quality cameras. Such a design allows users to define gestures easily, without having to spend significant time creating a training dataset. To this end, we introduce \textit{GRLib} - an open-source hand gesture detection and recognition Python library. The proposed solution utilizes a feed from an RGB, general-purpose camera. The obtained frames are first passed through an image augmentation pipeline, which goal is to increase the detection rates of \textit{MediaPipe Hands}. The landmarks obtained from \textit{MediaPipe Hands} are then passed to a standard classifier, such as K-Nearest Neighbours, in order to recognize the gesture. Both static and dynamic gestures are supported in the presented system. Furthermore, the system is capable of detecting and recognizing images in low quality video input. 

This paper is structured as follows. Section \ref{sec:static_methodology} introduces the method used for detecting and recognizing static gestures. Similarly, Section \ref{sec:dynamic_methodology} presents the approach for dynamic gestures. Section \ref{sec:results} contains an experimental evaluation of the library, including an ablation study for the different components of the system. Finally, the paper is concluded in Section \ref{sec:conclusion} along with some identified areas of future research.

%% file: sections/static_methodology.tex
\begin{figure}[h]
    \centering
    \includegraphics[scale=0.45]{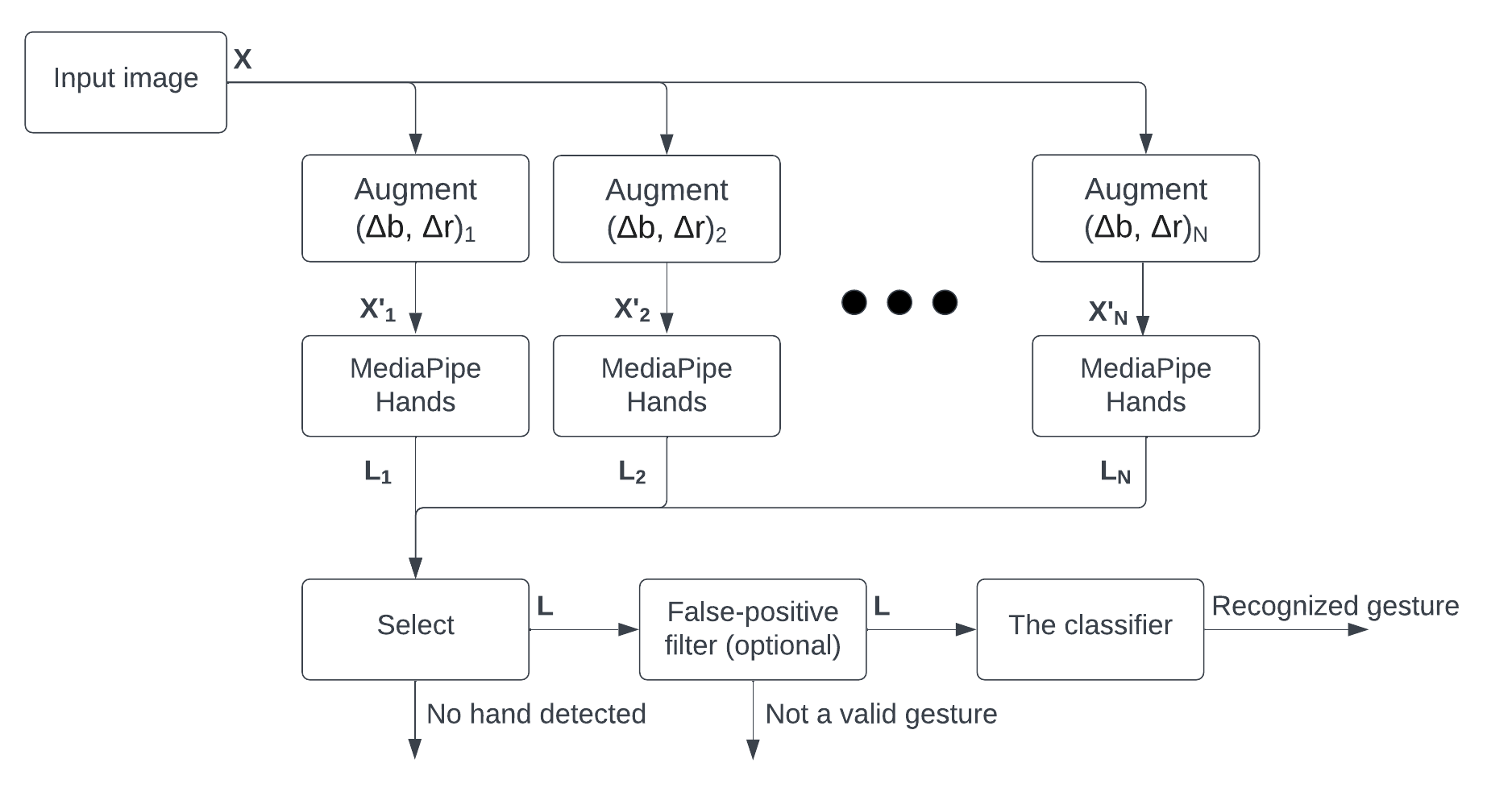}
    \caption{The architecture of the model used for recognizing static gestures. The input image is first passed through image augmentations, that change the brightness of the image or rotate it. \textit{MediaPipe Hands} is then applied to each augmented image separately, producing landmarks $L_i$. From $N$ landmarks, the first $L_i$ corresponding to at least one hand being detected is selected. The selected landmark $L$ is then passed through a false-positive filter and then, provided $L$ represents a valid gesture, it is passed to the classifier. The library will raise an exception if no hand was detected in any of the $N$ landmarks.}
    \label{fig:pipeline}
\end{figure}

Static gesture recognition involves detecting and recognizing gestures regardless of their position or motion. Given an input frame $X$, the goal is to predict whether the frame contains one of $C$ gesture classes or whether it contains only the background, i.e. when no valid gesture is shown. In this work, static gesture detection and classification are realized through four main components: image augmentations, \textit{MediaPipe Hands}, a false-positive filter, and a classifier. The architecture of the model is summarized in Figure \ref{fig:pipeline}. The following sections aim to cover each of the components of the system separately.

\subsection{Image augmentations}

Data augmentation is a technique that can be employed in scenarios where little training data is available \cite{few_shot_object_detection, few_shot_learning}. Augmentations that are commonly used in the computer vision domain include image flipping \cite{flipping_augmentation}, scaling \cite{scaling_augmentation}, rotation \cite{rotation_augmentation}, and changing brightness \cite{brightness_augmentation, few_shot_learning}. Typically, these augmentations are used to increase the amount of training data \cite{few_shot_object_detection}. In this paper, we propose the use of image augmentations in an attempt to increase the detection rates of \textit{MediaPipe Hands}. We reflect on the actual impact of using image augmentation on the detection rates in Section \ref{sec:results}.

In this work, image rotation and brightness changing are used to convert the single input frame $X$ into multiple augmented frames $X'_1, X'_2, \ldots, X'_n$. Each augmentation can be represented with a tuple $\left(\Delta b,~\Delta r\right)$, where $\Delta b$ is the update in brightness of the image and $\Delta r$ is the angle of rotation. Updating brightness is accomplished by converting each pixel $x_{RGB} \in X$ into its corresponding hue-saturation-value (HSV) representation $x_{HSV}$. Then the value component $x_V$ of $x_{HSV}$ is updated by $\Delta b$ using the formula:
\begin{equation}
    x_{V'} = \max (\min (x_{V} + \Delta b, 255), 0)
\end{equation}

The resulting $x_{HSV'}$ is converted back into the RGB representation $x_{RGB'}$. Applying this transformation to each pixel of $X$ results in the augmented image $X'_i$. Rotation is achieved using the \texttt{rotate\_bound} method from the \texttt{imutils} library \cite{imutils}.

\subsection{\textit{MediaPipe Hands}}

Similarly to \cite{gesture_control_computer_systems} and \cite{mediapipe_solutions} we use \textit{MediaPipe Hands} as the backbone of the system. Each of the augmented frames $X'_i$ is passed to MediaPipe to locate hand landmarks. Per each detected hand, \textit{MediaPipe Hands} returns 21 landmarks, each representing x, y, and z distances in meters to the landmark from the hand's approximate geometric center \cite{mediapipe}. On top of the landmarks, MediaPipe also returns `handedness` information, which indicates whether the landmarks correspond to the left or the right hand \cite{mediapipe}. The output of MediaPipe can thus be represented as a vector $L_i \in \mathbb{R}^{21 \cdot 3 \cdot n + n}$ with $n$ being the number of detected hands. The case when $n = 0$ is special as it corresponds to no hands being detected. In this case, we can represent the output of MediaPipe as $L_i = \emptyset$.

Special care has to be taken when gesture classes represented with 2 hands are allowed. It has been observed that sometimes even though 2 hands were shown, \textit{MediaPipe Hands} would only detect one, causing the vector $L_i$ to have different shapes. GRLib is thus parameterized by a value \texttt{num\_hands}, which represents the number of hands that should be detected. Additionally, the vector $L_i$ is constructed such that it first contains the landmarks of the left hand, then the right hand. Using this formulation, if only one hand was detected, the landmarks corresponding to the other hand are replaced with zeros and the handedness info is replaced with -1. Assuming at least one hand was found, the output vector $L_i$ will hence always have the size $21 \cdot 3 \cdot \texttt{num\_hands} + \texttt{num\_hands}$.

As the false-positive filter and the classifier only expect a single landmarks vector $L$ on input, a single vector $L_i$ needs to be selected. In this work, that is done by selecting the first landmarks vector where at least one hand was detected ($L_i \neq \emptyset$). We note, however, that different, perhaps more optimal, strategies could be implemented in an attempt to combine the found landmarks. If no hand was detected on any of the augmented frames, that is $\forall_i (L_i = \emptyset)$, the library raises an exception.

\subsection{False-positive filter}

The goal of the false-positive filter is to predict, given the landmarks $L$, whether the landmarks represent one of the $C$ gesture classes. This functionality is especially important in real-time classification, where it cannot be guaranteed that every shown gesture will represent one of the $C$ gestures. The false-positive filter is thus also optional and can be skipped if it is known that each image represents a valid gesture (for example during training).

Let $\mathcal{L}^c = \{L_1^c, L_2^c, \ldots, L_M^c\}$ be all of the training landmarks corresponding to the class $c$. For each class $c$, the false-positive filter then constructs a `representative sample` $R^c = \frac{1}{M}(L_1^c + L_2^c + \ldots + L_M^c)$. It should be noted that using the mean for the representative sample was found to work well in practice, however, the library could be extended to support different fusion strategies. For example, a median could be used to deal better with outliers. In order to then detect whether $L$ is a meaningful gesture, the false-positive filter computes the similarity between $L$ and each representative sample $R^c$. If the similarity is larger than a predefined threshold $\mu$ for any $R^c$, the landmark $L$ is passed to the classifier. Otherwise, the landmark is marked as a non-valid gesture. Currently, the library supports cosine and Euclidean similarity metrics. The value $\mu$ can be set by a user of the library and depends on how much variability in gestures should be permitted.

\subsection{The classifier}

The goal of the classifier is to predict, given the landmarks $L$,  one of the $C$ gesture classes. In \cite{gesture_control_computer_systems} and \cite{mediapipe_solutions} that is realized using a neural network. In this work, however, we offer the user freedom in choosing the classifier. Since the landmarks $L$ are guaranteed to be of the size $64 \cdot \texttt{num\_hands}$, any standard classifier would suffice for this task. The ablation study in Section \ref{sec:results} provides experimental results with using different classifiers.

%% file: sections/dynamic_meth.tex
Dynamic hand gestures (DHG) are gestures that consider not only the hand's shape but also its relative movements. Real sign languages primarily consist of dynamic gestures: for example, the 50 most important gestures of ASL are all dynamic \cite{handspeak}. Since people already use dynamic gestures to communicate with each other, recognizing them can be useful in a variety of fields, including human-computer interaction. In this section, we will explore the challenges and implemented approaches that were involved in recognizing dynamic hand gestures.

\subsection{Trajectories}
As described before, the hand's trajectory is one of the primary aspects of a dynamic gesture. Thus, handling it in different ways may have a great impact on the classifier's performance.

One relatively modern, yet common way of trajectory recognition is using Long Short-Term Memory (LSTM) \cite{lstm_dynamic} \cite{skeleton_lstm_dynamic_china}. With such an approach, a trained LSTM model receives a list of the hand's relative movements and predicts either a trajectory class or a gesture that corresponds to such trajectory. Although this approach can theoretically be scaled to recognize patterns of large size and complexity, it requires a significant amount of labeled training data \cite{NNs_bad_on_small_datasets} for both: the actual trajectories that have to be predicted, as well as pre-training. The focus of GRLib, however, is to allow few-shot or even single-shot learning. Consequently, the LSTM approach was deemed unsuitable for the task.

The sample efficient approach, that is implemented in GRLib was inspired by \cite{dynamic_directions_KSL}. The paper treats the trajectory as a series of general directions for each axis: positive movement, stationary, and negative movement. For example, the movement of a hand up and to the right, and then down and right would be encoded as ((+, +), (-, +)). Such a method has an important advantage - it generalizes the trajectory, stripping it of noise. Thus, the gesture's movement can be learned from a single sample.

\subsubsection{Key Frame Extraction}
To extract the described trajectory, a set amount of key frames is determined. Using them, it is possible to construct a trajectory that is not only of fixed length but also reduces the potential perceived jitters of the hand position that may occur as the hand is moving relative to the camera.

To extract representative key frames, we first remove outliers. A frame $i$ (denoted $f_i$) is considered an outlier if the following holds.
\begin{equation}
\label{eq:oulier_detection}
\operatorname{dist}(f_{i-1}, f_{i+1}) < \min[\operatorname{dist}(f_{i-1}, f_{i}), \operatorname{dist}(f_{i}, f_{i+1})],
\end{equation}
where $\operatorname{dist}(f_i, f_j)$ is the distance between the hand in frame $i$ to hand in frame $j$. The key frames are determined by evenly distributing the cumulative distance covered by the hand movement between two sequential frames, always including the first and the last frames.

\subsection{Gesture start detection}
Detecting dynamic gestures in real time poses an additional challenge: without special markers, it is impossible to know that a gesture has started, before it is complete, as the hand may perform only the beginning of the sequence. A possible solution to this problem is through the use of gesture start markers, as presented by \textcite{gesture_control_computer_systems}. The requirement for the markers imposes a significant limitation: apart from performing a gesture, the user has to use another means of input to indicate its start and end. This can interrupt the fluidity of the interaction and further complicate intuitive usage. It was therefore decided to adapt the library to work without the markers.

\subsubsection{Prediction Candidates}
\begin{figure}[h]
    \centering
    \includegraphics[scale=0.45]{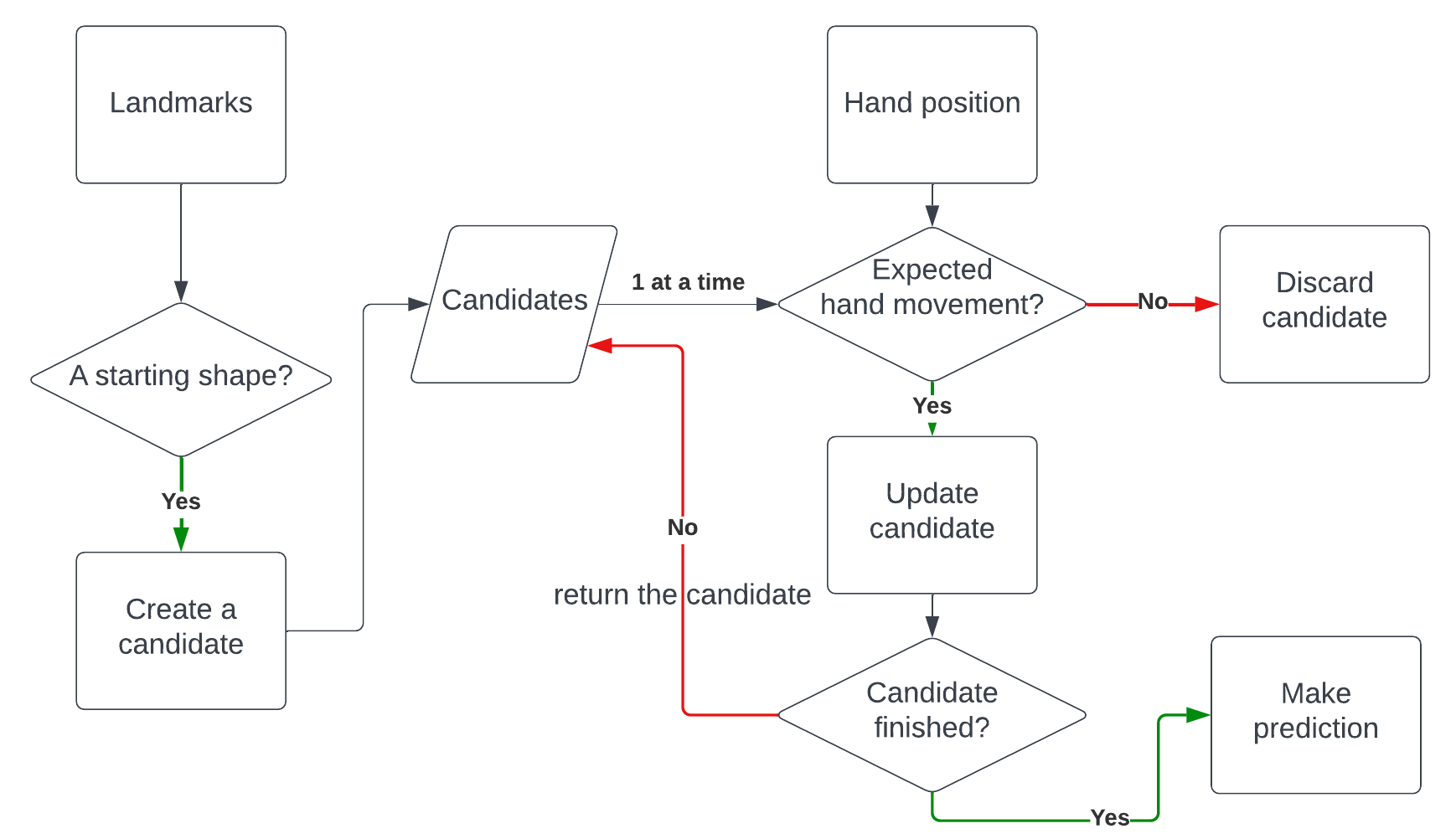}
    \caption{Schematic representation of the candidate update process. First, a classifier is used to determine whether the landmarks in the current frame represent the starting shape of a gesture. If so, a new \textit{candidate} is created for each matching gesture class. Then, for each \textit{candidate}, it is checked if the current hand position aligns with the expected movement for the class trajectory. If so, the \textit{candidate's} trajectory is updated, and in case the trajectory is complete, the prediction is made. Otherwise, the \textit{candidate} gets put back into the \textit{candidate} storage.}
    
    \label{fig:candidates}
\end{figure}

The chosen approach keeps a list of so-called \textit{candidates}: known gestures that can currently be shown, and their respective progress. 
Candidates are added to the dedicated list when a start of any known dynamic gesture is possible. Specifically, when the hand's shape matches the starting shape of a dynamic gesture. If the hand shape matches the starting shape of multiple classes, a candidate is created for every one of them. Every set amount of frames, each of the current candidates is checked: the new position of the hand is compared with the previous position, and if the hand's actual trajectory aligns with the trajectory expected by a candidate, the latter is kept and updated. Otherwise, the candidate is dropped as irrelevant. When one of the candidates successfully reaches its end, the prediction is made and the others are cleared. A diagram of the algorithm for one frame is presented in Figure \ref{fig:candidates}. As a result, the algorithm continuously evaluates possible paths to a complete dynamic gesture.

%% file: sections/experiments.tex
\section{Results}
\label{sec:results}

The following section assesses the ability of the library to perform gesture detection and recognition. The topics of static and dynamic gestures are handled separately. The analysis for static gestures is done through the means of an ablation study that shows the impact of each component of the system on the overall performance. We conducted an ablation study for evaluation because the absence of similar systems in the literature makes direct comparisons challenging \cite{gesture_control_computer_systems}. In the literature, only two similar libraries have been identified: \textit{MediaPipe Solutions} \cite{mediapipe_solutions} and Gestop \cite{gesture_control_computer_systems}. Gestop does not however support existing datasets as a source of the training data. On top of the ablation study, we also compare the performance of GRLib with \textit{MediaPipe Solutions}.

\subsection{Static gestures}

In this section, we explore the library's capabilities in detecting and recognizing static gestures. This is done through the means of an ablation study. Additionally, the performance of the library is measured against \textit{MediaPipe Solutions}.

\textbf{Datasets.} Three datasets of varying complexity were chosen for the assessment of static HGR capabilities. First, the American Sign Language dataset \cite{ASL_dataset}, which contains 3,000 images per English letter, all of which share a similar background. The second dataset is a subsampled version of HaGRID \cite{hagrid} with 100 images per class. Gestures for 18 classes are performed by different people in various environments. The third dataset is taken from the Kenyan Sign Language Classification Challenge \cite{KSL_challenge}, it contains 9 static classes of gestures involving one or two hands. The complexity is increased by the quality of the images: on some of them, the hands are partly blurred out, while fingers and/or forearms may be out of frame. Additionally, the Kenyan competition dataset is the only tested dataset where gesture classes requiring two hands are present.

\textbf{Evaluation metric.} As in this work an emphasis is put on the detection rates, a metric combining both detection rate and classification accuracy needs to be defined. Let the training dataset have $N$ images, of which in only $M$ a gesture was detected. The detection rate is then defined as $r = \frac{M}{N}$. Given the classification accuracy $c$, the final metric $m$ is calculated as $m = c \cdot r = c \cdot \frac{M}{N}$. Thus, a better model is expected to have a higher value of the metric $m$.

\subsubsection{Image augmentations}
\label{sec:image_augmentations_study}

In this section, the impact of using image augmentations is shown on the detection rate $r$ and the combined metric $m$. To show the performance of the model in low-data regimes, the model is trained through an $n$-shot learning procedure, where $n$ images per each class are subsampled from the training dataset at random. The model is then evaluated on the remaining images from the training dataset. The procedure is repeated 10 times for each $n$, at each time with a different, randomly subsampled set. 

\begin{table}[t]
    \centering
    \caption{Four different settings for the image augmentations that were used for testing. Setting 1 is the baseline for comparison when no augmentations are done. Setting 2 modifies only the brightness of the image while setting 3 only the rotation of the image. Setting 4 is a combination of both brightness and rotation.}
    \label{tab:augmentations_overview}
    \begin{tabular}{|c|c|c|c|c|c|c|c|}
        \hline
        \multirow{2}{*}{Settings} & \multicolumn{7}{c|}{Stages $(\Delta b, \Delta r)$}\\
        \cline{2-8}
         & Stage 1. & Stage 2. & Stage 3. & Stage 4. & Stage 5. & Stage 6. & Stage 7.\\
        \hline
        Setting 1. & (0, 0) & & & & & &\\
        Setting 2. & (0, 0) & (30, 0) & (60, 0) & & & &\\
        Setting 3. & (0, 0) & (0, -15) & (0, 15) & (0, -30) & (0, 30) & &\\
        Setting 4. & (0, 0) & (30, 0) & (60, 0) & (0, -15) & (0, 15) & (0, -30) & (0, 30) \\
        \hline
    \end{tabular}
\end{table}

\begin{figure*}[t]
    \centering
    \begin{subfigure}{0.32\textwidth}
        \includegraphics[width=\textwidth]{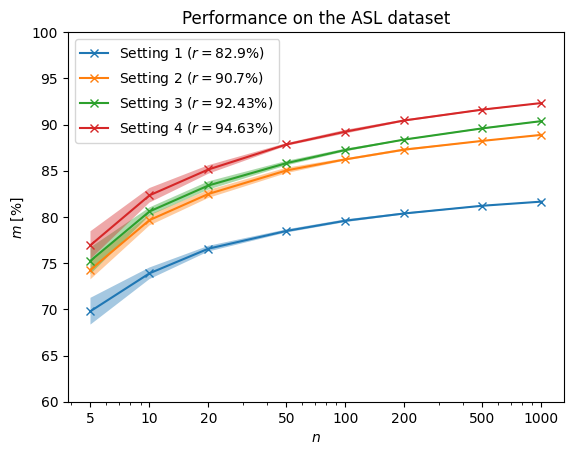}
    \end{subfigure}
    \begin{subfigure}{0.32\textwidth}
        \includegraphics[width=\textwidth]{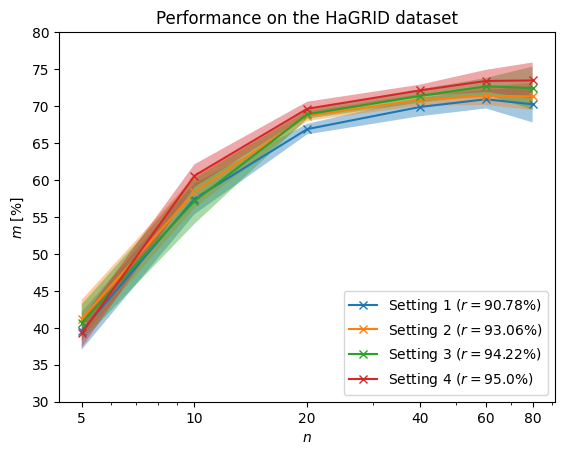}
    \end{subfigure}
    \begin{subfigure}{0.32\textwidth}
        \includegraphics[width=\textwidth]{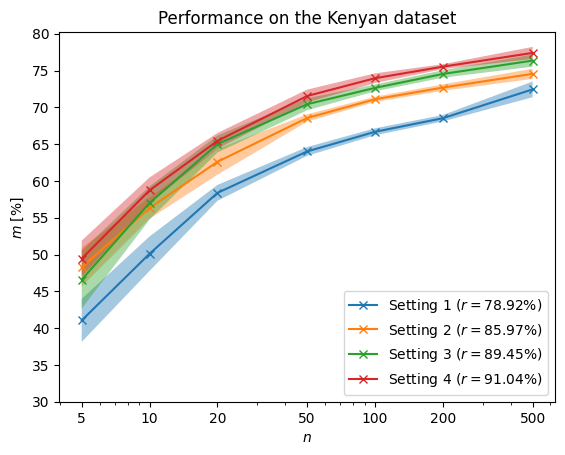}
    \end{subfigure}
    \caption{Performance of GRLib on ASL \cite{ASL_dataset}, HaGRID \cite{hagrid}, and Kenyan \cite{KSL_challenge} datasets. The x-axis represents the number $n$, which is the number of training samples available for each class. The y-axis represents the metric $m$, obtained by multiplying the accuracy score $a$ with the recognition rate $r$. The recognition rates for each augmentation setting are shown in brackets in the legend. The width of each line corresponds to two standard deviations.}
    \label{fig:results_augmentations_study}
\end{figure*}

\begin{figure*}[ht]
    \centering
    \begin{subfigure}{0.32\textwidth}
        \includegraphics[width=\textwidth]{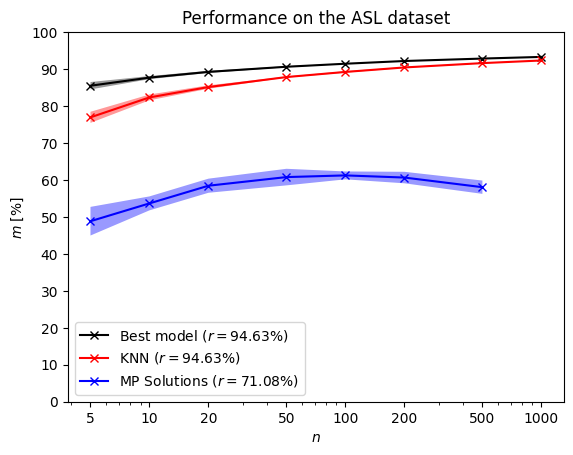}
    \end{subfigure}
    \begin{subfigure}{0.32\textwidth}
        \includegraphics[width=\textwidth]{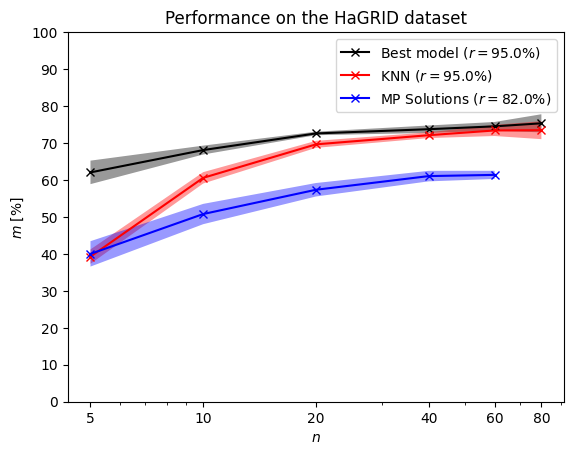}
    \end{subfigure}
    \begin{subfigure}{0.32\textwidth}
        \includegraphics[width=\textwidth]{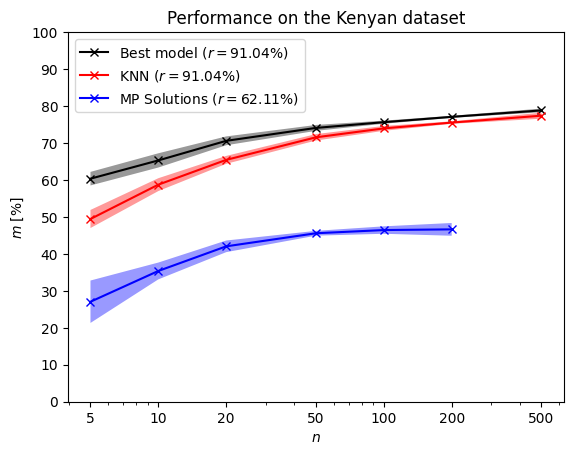}
    \end{subfigure}
    \caption{Performance of GRLib on ASL \cite{ASL_dataset}, HaGRID \cite{hagrid}, and Kenyan \cite{KSL_challenge} datasets using different classifiers. The x-axis represents the number $n$, which is the number of training samples available for each class. The y-axis represents the metric $m$, obtained by multiplying the accuracy score $a$ with the recognition rate $r$. Performance of both, the best model found, and the K-Nearest Neighbours classifier used in Section \ref{sec:image_augmentations_study} are shown. Furthermore, the performance of \textit{MediaPipe Solutions} \cite{mediapipe_solutions} is reported. The width of each line corresponds to two standard deviations obtained by repeating the experiment 10 times.}
    \label{fig:results_classifier_study}
\end{figure*}

The testing procedure is done on the three datasets, using 4 different image augmentation settings. For a baseline comparison, one of the settings contains no augmentations. The other three settings modify only the brightness, only the rotation, or a combination of both. The exact overview of conditions can be found in Table \ref{tab:augmentations_overview}. As for the classifier, K-Nearest Neighbours (KNN) with $k=5$ was chosen. The inputs to the classifier are only the detected landmarks, without the handedness info. This was done as the dimensionality of the handedness info is orders of magnitude larger than the landmarks. Figure \ref{fig:results_augmentations_study} shows the results of the experiment.

The figure demonstrates the number of training samples per class $n$ on the x-axis and the average metric $m$ on the y-axis. As the sizes of each dataset are different, different values of $n$ were chosen for each one of them. The difference between the top and the bottom of each line corresponds to two standard deviations obtained by repeating the experiment 10 times, which hence represents the 68\% confidence interval. The recognition rates $r$ for each of the image augmentations settings are shown in brackets in the legend. As can be observed from the figure, introducing image augmentations increases the recognition rates $r$ with respect to the baseline on all three datasets. Consequently, this usually also leads to an increase in the evaluation metric $m$, although the results for the HaGRID dataset are inconclusive due to the large variance in the results.

\subsection{The classifier}

The following section aims to show how the performance of the library changes with different classifiers. Similarly to the experiments on image augmentations, the models are trained on $n$ images per class and tested on the remainder. The procedure is repeated 10 times each time randomly sampling the $n$ images. Training and testing are done using the landmarks obtained by using Setting 4 for the image augmentations. This setting was chosen as it exhibited the highest recognition rates for all three datasets. The best model was then the one that obtained the highest average metric $m$ over the 10 repetitions over all of the values of $n$. To report the final performance, the best-found model was then trained and evaluated from scratch for each $n$, repeating the experiments 10 times. 

Four different classifiers are evaluated using this procedure: K-Nearest Neighbours (KNN), Support Vector Machine (SVC), Decision Tree Classifier, and Logistic Regression. Furthermore, for each model, the best combination of its parameters would be attempted to be found. The exact list of parameters searched for and the corresponding experimental result can be found in the repository of the project. Additionally, we report the performance of MediaPipe Solutions using the same experimental setup. This is done to compare the results using different techniques. The best-found classifier is also further compared to the K-Nearest Neighbour classifier used in the previous section. The comparison can be seen in Figure \ref{fig:results_classifier_study}.

In all 3 cases, the SVC classifier with the radial basis function (RBF) kernel was selected as the best model. For both the ASL and HaGRID datasets, the best model had the parameter $C=20$. The model for the Kenyan dataset had $C=5$. As can be observed from Figure \ref{fig:results_classifier_study}, the choice of the model influences the classification performance, especially in low-data regimes. This is most visible in the HaGRID dataset for $n = 5$, where the SVC model obtained $m = 0.62$ compared to KNN's $m = 0.39$. This is perhaps expected as the support vector machine classifier attempts to separate the classes as much as possible.

Furthermore, the performance of the model is compared on the three datasets against \textit{MediaPipe Solutions} \cite{mediapipe_solutions}. As can be observed, the recognition rates, visible in the legend in Figure \ref{fig:results_classifier_study}, are much lower for \textit{MediaPipe Solutions}. The lower recognition rates also made it impossible to evaluate \textit{MediaPipe Solutions} on the largest values of $n$. There would not be enough images to sample 1000, 80, and 500 per class from the three datasets. Overall, it can be observed that \textit{GRLib} outperforms \textit{MediaPipe Solutions} on all tested datasets.

\subsection{Dynamic gestures}

This section outlines the library's capability for real-time identification of dynamic gestures. Our algorithm is tailored for both sample efficiency and online operation. During inference, it continuously processes incoming frames without requiring markers for the start and end of gestures, thus generating a continuous stream of predictions.

To test the performance of the proposed solution, we create a custom dynamic gesture dataset. The dataset contains five training samples for 9 gesture classes and evaluates the model in an online setting. The classes were chosen in the following way: 
\begin{itemize}
    \item 2 gestures (palm-left and palm-right) differ exclusively in the direction of the hand motion.
    \item 2 gestures (right-fist-to-palm and right-palm-to-fist) share a simple trajectory with other gestures, but can be distinguished by the start and end hand shapes.
    \item 3 gestures (fist-up, forward, backward) share the start hand shape but have different end shapes and go in 3 different directions, including the depth dimension.
    \item 2 gestures (palm-circle, palm-infinity) have a lengthy and complex trajectory with a constant hand shape - open palm.
\end{itemize}
With this dataset, we aim to highlight the most important characteristics of the DHG recognition algorithm.

For inference, we use multiple sequences of various nature. One of the sequences repeats simple gestures three time each, 3 others present a sequence of gestures one flowing into another. In about half of the sequences, gestures are interleaved with background actions that do not correspond to any gesture class. All sequences contain more than 3 presented gestures to simulate the online evaluation, rather than feeding one gesture at a time.

\begin{figure*}[ht]
    \centering
    \begin{subfigure}{0.5\textwidth}
        \includegraphics[width=\textwidth]{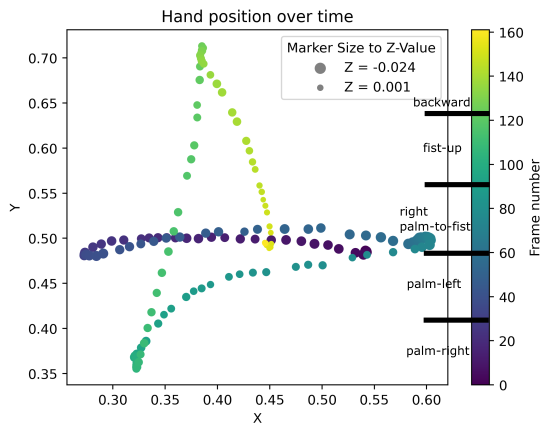}
    \end{subfigure}
    \caption{The 3d hand trajectory from one of the flowing sequences.  The color indicates time, and the marker size indicates the z value (closer to the camera is bigger). Near the frame number, the ground truth classes are labeled.}
    \label{fig:online_trajectories}
\end{figure*}

Example hand movements from the dataset are presented in Figure \ref{fig:online_trajectories}. We publish the dataset alongside the library.

The library has shown good performance with a limited amount of training data. The overall gesture recognition recall was 79\%, despite the challenges associated with the sparsity of the training data. However, we observe a tendency of the system to identify the same gesture more than once consecutively. This happens with gestures that have a trivial trajectory - moving the hand in one direction. Then, as the hand moves, the library predicts this gesture. In case the hand continues to move in the same direction after the prediction, the prediction can be repeated.

Another experienced problem was with gestures containing a relatively complex trajectory - circle and infinity are not recognized at all. Due to the way the trajectories are encoded - with the key frames and directional quantization, the complex trajectories are inherently difficult to correctly encode, let alone recognize.

We also noticed that the depth estimation of MediaPipe is problematic, leading to the gestures "forward" and "backward" being frequently falsely detected. Therefore, we reran the experiments without introducing these gestures. Not accounting for repetitions, we observe 100\% detection recall on gestures other than "circle" and "infinity", and 0\% recall on these two. As for precision, we get about 60\% overall, but a much higher about 95\% when not considering additional predictions of the same correct gesture.

Not accounting for landmark extraction, dynamic gesture detection is able to run at $2083\pm13$ frames per second on a laptop with 2.6 GHz 6-Core Intel Core i7. Thus, it is demonstrated to be highly suitable for real-time applications.

Despite the aforementioned double-detection issue, the results indicate that our hand gesture recognition library shows promising capabilities for real-time, online applications, even with a limited number of training samples.

%% file: sections/conclusion_future_work.tex
\section{Conclusion and Future Work}
\label{sec:conclusion}

Hand gesture recognition (HGR) systems provide a natural way of human-computer interaction. Scientific studies that try to improve robustness and accuracy of these systems vary from skin colour detection \cite{colour_based_comparison, colour_based_survey} to applying machine learning for gesture recognition \cite{cnn_hgr, rcnn_hgr, vision_based_review}. However, existing systems suffer in scenarios where the background is complex or where the lighting is imperfect \cite{difficulties_vision_based_systems, robust_real_time, robust_kinect}. In this paper, we identified that there is a lack of publicly available libraries that are able to work under difficult conditions and that allow users to define their own gestures, which in turn can be trained on prerecorded and publicly available gesture data. \cite{gesture_control_computer_systems}.

Work by Krishna and Sinha \cite{gesture_control_computer_systems} introduced \textit{Gestop}, a library that allows the user to define their own gestures to control computer systems. Both static and dynamic gestures are supported by the system. Their library, however, does not support gestures to be defined from already existing datasets. \textit{MediaPipe Solution} offers a similar system, where the user can also define their own gestures. This package however does not support dynamic gestures.

In this work, we presented \textit{GRLib} - an open-source, Python hand gesture recognition library. The proposed solution uses \textit{MediaPipe Hands} to detect landmarks, which are then passed to a classifier chosen by the user of the library. The landmark detection is further enhanced by means of image augmentations. A false-positive filter is introduced, which helps in scenarios where it can not be guaranteed that a valid hand gesture is shown. A method for recognizing dynamic gestures is also implemented.

The system is evaluated on three real-world datasets: American Sign Language (ASL) \cite{ASL_dataset}, Kenyan \cite{KSL_challenge}, and HaGRID \cite{hagrid}. We find that the image augmentations greatly increase the detection rates of \textit{MediaPipe Hands} on all three datasets. Furthermore, we find that the choice of the classifier improves the performance of the library in a few-shot learning scenario. Moreover, it is also found that the system outperforms \textit{MediaPipe Solutions} on all three datasets.

In the future, a user test could be performed. The participants could be instructed to apply the library to different problems including static and dynamic gestures. Such experiments could for example include building a desktop application that uses a laptop camera or controlling robot movements with gestures. Dependent measures may vary from objective measures, for example task completion time, to subjective questionnaires regarding ease of use. This feedback could be used to improve the usability of the proposed library and augment it with useful components. Additionally, the functional component of the library could be improved. More image augmentations, such as changing contrast and saturation \cite{augmentations_for_CNNs} could be included. The library can also be extended to work with more than 2-hand gestures.

Dynamic gesture recognition can also be improved in the future. We have observed that the current implementation struggles with long or complex trajectories. This is likely caused by the way key frame extraction is done. For example, for an 8-shaped trajectory, it is unlikely that a small number of key frames will approximate the true hand movement well. A different approach towards selecting key frames may be able to mitigate this issue. For example, an algorithm may attempt to put the key frames in the ``corners" of a trajectory. Additionally, during inference, the candidate update process is done in equal intervals, while it could take into account past hand movement to update candidates at better moments.